\documentclass{article}
\usepackage{ijcai23}

\usepackage{times}
\usepackage{soul}
\usepackage{url}
\usepackage[hidelinks]{hyperref}
\usepackage[utf8]{inputenc}
\usepackage[small]{caption}
\usepackage{graphicx}
\usepackage{amsmath}
\usepackage{amsthm}
\usepackage{booktabs}
\usepackage{algorithm}
\usepackage{algorithmic}
\usepackage[switch]{lineno}

\newtheorem{example}{Example}

\usepackage{tikz}
\usetikzlibrary{shapes,decorations,arrows,calc,arrows.meta,fit,positioning}
\tikzset{
    -Latex,auto,node distance =1 cm and 1 cm,semithick,
    state/.style ={ellipse, draw, minimum width = 0.7 cm},
    point/.style = {circle, draw, inner sep=0.04cm,fill,node contents={}},
    bidirected/.style={Latex-Latex,dashed},
    el/.style = {inner sep=2pt, align=left, sloped}
}

\newtheorem{definition}{Definition}
\newtheorem{scenario}{Scenario}

\usepackage{subcaption}

\title{A New Paradigm for Counterfactual Reasoning in Fairness and Recourse}

\author{
Lucius E.J. Bynum$^1$
\and
Joshua R. Loftus$^2$\And
Julia Stoyanovich$^1$
\affiliations
$^1$New York University, Center for Data Science\\
$^2$London School of Economics, Department of Statistics\\
\emails
lucius@nyu.edu,
J.R.Loftus@lse.ac.uk,
stoyanovich@nyu.edu
}

\begin{document}

\maketitle

\begin{abstract}
    Counterfactuals and counterfactual reasoning underpin numerous techniques for auditing and understanding artificial intelligence (AI) systems. 
    The traditional paradigm for counterfactual reasoning in this literature is the \emph{interventional counterfactual}, where hypothetical interventions are imagined and simulated.
    For this reason, the starting point for causal reasoning about legal protections and demographic data in AI is an imagined intervention on a legally-protected characteristic, such as ethnicity, race, gender, disability, age, etc.
    We ask, for example, what would have happened had your race been different?
    An inherent limitation of this paradigm is that some demographic interventions --- like interventions on race --- may not translate into the formalisms of interventional counterfactuals.
    In this work, we explore a new paradigm based instead on the \emph{backtracking counterfactual}, where rather than imagine hypothetical interventions on legally-protected characteristics, we imagine \emph{alternate initial conditions} while holding these characteristics fixed.
    We ask instead, \emph{what would explain a counterfactual outcome for you as you actually are or could be?}
    This alternate framework allows us to address many of the same social concerns, but to do so while asking fundamentally different questions that do not rely on demographic interventions.
\end{abstract}

\section{Introduction}

Counterfactual reasoning plays a pivotal role in many modern techniques for auditing and understanding machine learning models and artificial intelligence (AI) systems. Traditionally, counterfactual discrimination criteria ask the following question:
\emph{How would this decision-making system behave if, counterfactually, an individual had a different value of their legally-protected characteristic?} For example, would a given person have more likely received a financial loan if they were not in a minority group? Depending on the setting, such behavior from a decision-making system can be seen as evidence of discrimination \cite{kilbertus2017avoiding,Nabi2017FairIO,Loftus2018CausalRF}.

While causal reasoning, and counterfactual reasoning in particular, open important avenues of analysis, the primary starting point for these analyses --- an imagined intervention on a legally-protected characteristic --- is also a limitation. Think, for example, about analyzing US Census data, which has features like birthplace, racial category, socioeconomic status, and ethnicity. To imagine and simulate interventions in such a setting can require us to have unrealistically precise and modular definitions of complex social categories like race: these features are often used to \emph{define} each other, and cannot easily be separated into distinct interventions.\footnote{We can state this formally in terms of \emph{modularity} (see \S~\ref{sec:background}).} Further, it can be unclear what such interventions actually refer to.
If someone both (1) cares about making more equitable decisions, and (2) believes that causal structure plays an important role in governing the context around a decision-making system, they are left with few tools for counterfactual reasoning that do not suffer from this limitation.

\begin{figure*}
    \centering
    \includegraphics[width=\textwidth]{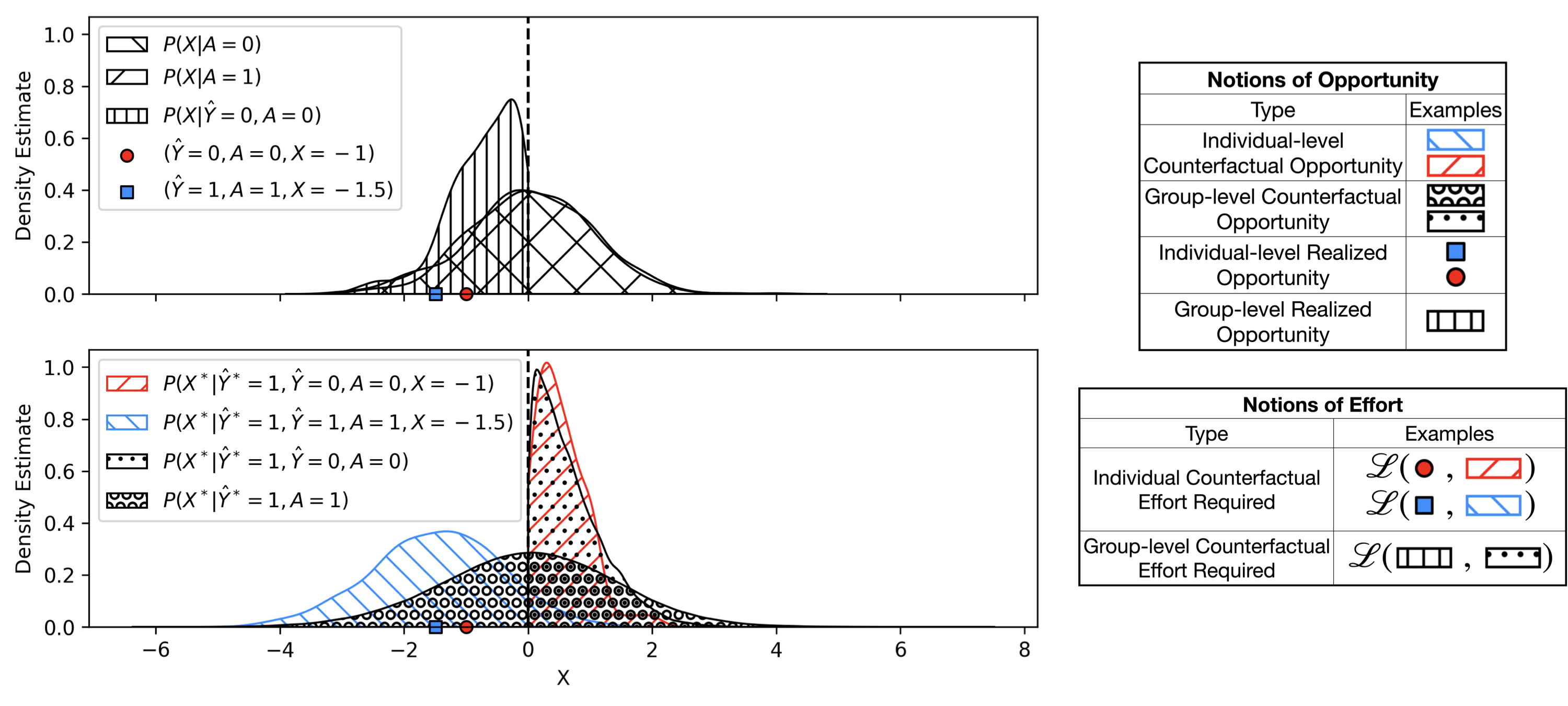}
    \caption{A visual example of several backtracking counterfactual quantities to consider for algorithmic fairness and recourse (labeled on the right). Based on Example~\ref{ex:running_example}, consider covariate $X \sim \mathcal{N}(0, 1)$, protected attribute $A \sim \text{Bern}(0.5)$, and predictor $\widehat{Y}=A \lor (X > 0)$. A corresponding DAG is shown in Figure~\ref{fig:dag_backtracking_cf}. With backtracking conditional $P_B(U^* \mid U) = \left(U^*_A = U_A, U^*_X = \mathcal{N}(U_X, 1), U^*_Y = 0\right)$ and an appropriate cost function $\mathcal{L}$, we can use Algorithm~\ref{alg:backtracking_counterfactual_sampling} to estimate the various notions of opportunity and effort shown above for both individuals and groups. This process is formally defined in Sections~\ref{sec:opportunity}~\&~\ref{sec:effort}.}
    \label{fig:main_figure}
\end{figure*}

In this paper, we propose a new paradigm for counterfactual reasoning in algorithmic fairness and recourse that alleviates this problem. This paradigm starts with a different counterfactual question: \emph{What would explain a counterfactual outcome for an individual as they actually are or could be?} Starting with this question allows us to still reason counterfactually about fairness and recourse, but to do so without relying on hypothetical demographic interventions. 

Underpinning these two different counterfactual questions are two different semantics for computing counterfactuals. The \emph{interventional counterfactual}, where hypothetical interventions are imagined and simulated, is the dominant semantics used for computing counterfactuals across computer science literature. In this context, counterfactual discrimination criteria have primarily been formalized by imagining hypothetical interventions on demographic variables or their proxies \cite{Kusner2017CounterfactualF,Chiappa2018PathSpecificCF,Wu2019PCFairnessAU,Pleko2022CausalFA,kilbertus2017avoiding,Zhang2018FairnessID,von2022fairness,Ehyaei2023RobustnessIF,causal_fairness_field_guide}. Recent work has formalized an alternate semantics for computing counterfactuals: the \emph{backtracking counterfactual} answers counterfactual queries by imagining changes to initial conditions instead of simulating hypothetical interventions \cite{von2023backtracking}. Just as the interventional counterfactual is a natural fit for reasoning about demographic interventions, the backtracking counterfactual is a natural fit for explaining counterfactual outcomes, and thus, a natural fit for our alternate counterfactual question.

\paragraph{Contribution.} In what follows, we define new technical notions of counterfactual opportunity and effort, and introduce several new counterfactual discrimination criteria. 
These are, to our knowledge, the first counterfactual-based fairness criteria that depart from the need to consider intervention on legally-protected characteristics, while still allowing us to simultaneously (1) consider legally-protected characteristics, (2) make use of causal relationships between variables, and (3) consider fairness at an individual or a group level. We also develop a simple algorithm for sampling backtracking counterfactuals and use it to implement several of the proposed criteria on real and simulated data. In effect, this paper lays the foundation for a potentially large body of work. We view this as the first step in a wider research agenda that charts a course for future work across algorithmic fairness and recourse and, more broadly, develops a new approach to counterfactual analysis for demographic data.

The core idea behind the counterfactual discrimination criteria we introduce in this work is to quantify unfairness for an individual or group in terms of their \emph{opportunity} or their \emph{required effort} --- factually and counterfactually. As a running example, consider an algorithm that is used to assist in hiring decisions or recruitment. 

\begin{example}[Discrimination in Hiring]\label{ex:running_example}
     Let binary variable $A$ represent whether or not a person is in a majority group, binary $Q$ represent whether or not some job-related qualifications $X$ exceed a desired threshold, and binary $\widehat{Y}$ represent whether or not they receive a job offer. Consider the following discriminatory hiring practice: $\widehat{Y} = Q \lor A$, or in other words, a person is offered the job either if they are in the majority group or if they have the relevant qualifications. 
\end{example}

 At the level of intuition, in Example~\ref{ex:running_example} we can say that a person in the minority group with $A=0$ has only one opportunity to get the job --- they need $Q=1$ --- whereas people in the majority group with $A=1$ have more opportunity to get the job, including both $Q=0$ and $Q=1$. \footnote{For the reader concerned about `fairness through blindness,' the operative question is this: \emph{Is the variable $Q$ measuring only morally relevant qualifications, or is it also measuring different levels of access to resources?}}
 See Figure~\ref{fig:main_figure} for a visualization of several ways we can use backtracking counterfactuals to represent opportunity and effort in Example~\ref{ex:running_example}. In the remainder of this paper, we formalize how to compute the quantities in Figure~\ref{fig:main_figure}, both for this example and for arbitrary settings with possibly larger sets of variables and more complicated relationships between them.

\section{Mathematical Preliminaries}\label{sec:background}

In this section, we provide the necessary mathematical background to define our counterfactual discrimination criteria, with structural causal models and interventional counterfactuals following \cite{Pearl2000CausalityMR} and backtracking counterfactual semantics and notation from \cite{von2023backtracking}.

\paragraph{Causal models.}
We define a \emph{causal model} as a tuple $\mathcal{M} = (V, U, F)$. In this tuple, $V$ is a set of observed variables, $U$ a set of unobserved (exogenous) variables, and $F$ a set of functions $\{f_i\}_{i=1}^{|V|}$ for each $V_i \in V$ such that $V_i = f_i(P_i, U_{P_i})$ where $P_i \subseteq V \setminus \{ V_i\}$ represents the causal parents of $V_i$ and $U_{P_i} \subseteq U$. A causal model can be pictorially represented as a directed acyclic graph (DAG) with nodes for $U, V$ and directed edges for $F$. A \emph{probabilistic causal model} $(\mathcal{M}, P(U))$ adds distribution $P(U)$ over the unobserved variables.

\paragraph{Interventions and counterfactuals.}
We define an atomic \emph{intervention} on variable $V_i$ as the substitution of equation $V_i = f_i(P_i, U_{P_i})$ with a particular value $V_i = v$, simulating the forced setting of $V_i$ to $v$ by an external agent, commonly notated $\text{do}(V_i = v)$. A \emph{submodel} $\mathcal{M}_x$ for realization $x$ of $X \subseteq V$ is the model $\mathcal{M}$ after intervention $\text{do}(X=x)$. Given a probabilistic causal model, we can derive from $(\mathcal{M}_x, P(U))$ the distribution of any subset of variables following intervention $\text{do}(X=x)$. Using instead a distribution over the exogenous variables that is specific to a particular context or individual, the same mechanics allow us to define \emph{interventional counterfactuals} to model alternate possible outcomes after interventions in a specific context. Following \cite{von2023backtracking,Balke1994ProbabilisticEO}, we use an asterisk to denote counterfactual versions $V^*$ of variables $V$. Counterfactual variable $Y^*$ given a factual observation $z$ and intervention $\text{do}(X=x^*)$ ($X, Y, Z \subseteq V$) can be computed via a three-step procedure often referred to as `abduction, action, prediction.' Abduction uses observed evidence to obtain $P(U \mid z)$ from $P(U)$. Action performs intervention $\text{do}(X=x^*)$ for counterfactual realization $x^*$ to obtain $\mathcal{M}_{x^*}$. Prediction computes the probability of $Y^*$ from $(\mathcal{M}_{x^*}, P(U \mid z))$.

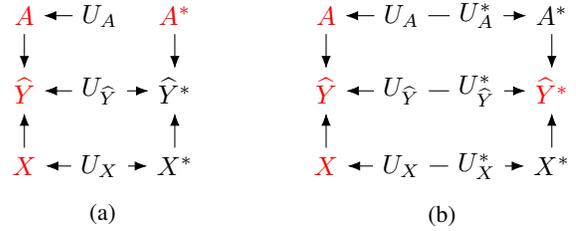
\begin{figure}
    \begin{subfigure}[c]{0.2\textwidth}
        \centering
        \begin{tikzpicture}
            \node (1) at (-1, 1) {\textcolor{red}{$A$}};
            \node (2) at (-1, 0) {\textcolor{red}{$\widehat{Y}$}};
            \node (3) at (-1, -1) {\textcolor{red}{$X$}};

            \node (4) at (0, 1) {$U_A$};
            \node (5) at (0, 0) {$U_{\widehat{Y}}$};
            \node (6) at (0, -1) {$U_X$};

            \node (7) at (1, 1) {\textcolor{red}{$A^*$}};
            \node (8) at (1, 0) {$\widehat{Y}^*$};
            \node (9) at (1, -1) {$X^*$};
        
            \path (1) edge (2);
            \path (3) edge (2);

            \path (4) edge (1);
            \path (5) edge (2);
            \path (6) edge (3);
            \path (5) edge (8);
            \path (6) edge (9);

            \path (7) edge (8);
            \path (9) edge (8);
        \end{tikzpicture}
        \caption{}
        \label{fig:dag_regular_cf}
    \end{subfigure}
    \begin{subfigure}[c]{0.3\textwidth}
        \centering
        \begin{tikzpicture}
            \node (1) at (-1, 1) {\textcolor{red}{$A$}};
            \node (2) at (-1, 0) {\textcolor{red}{$\widehat{Y}$}};
            \node (3) at (-1, -1) {\textcolor{red}{$X$}};

            \node (4) at (0, 1) {$U_A$};
            \node (5) at (0, 0) {$U_{\widehat{Y}}$};
            \node (6) at (0, -1) {$U_X$};

            \node (7) at (2, 1) {$A^*$};
            \node (8) at (2, 0) {\textcolor{red}{$\widehat{Y}^*$}};
            \node (9) at (2, -1) {$X^*$};

            \node (10) at (1, 1) {$U^*_A$};
            \node (11) at (1, 0) {$U^*_{\widehat{Y}}$};
            \node (12) at (1, -1) {$U^*_X$};
        
            \path (1) edge (2);
            \path (3) edge (2);

            \path (4) edge (1);
            \path (5) edge (2);
            \path (6) edge (3);

            \draw[-] (4) edge (10);
            \draw[-] (5) edge (11);
            \draw[-] (6) edge (12);

            \path (7) edge (8);
            \path (9) edge (8);

            \path (10) edge (7);
            \path (11) edge (8);
            \path (12) edge (9);
        \end{tikzpicture}
        \caption{}
        \label{fig:dag_backtracking_cf}
    \end{subfigure}
    \caption{Graphical models for two counterfactual fairness problems with protected attribute $A$, features $X$, and classification outcome $\widehat{Y}$. (a) Interventional approach with intervention $\text{do}(A=A^*)$. (b) Backtracking approach with counterfactual observation $\widehat{Y}^*$. Red nodes are known. Based on Figure~2 in \protect\cite{von2023backtracking}.}
    \label{fig:dags_cf_example}
\end{figure}

\paragraph{Backtracking counterfactuals.} Backtracking counterfactuals suggest instead a semantics of counterfactuals that \emph{leaves all of the causal mechanisms unchanged}. Instead of external actions being simulated via intervention, alternate possible outcomes are explained via changes to the values of exogenous variables. With unchanged mechanisms, there are many possible exogenous changes that can explain counterfactual observations. Backtracking counterfactuals answer probabilistic queries about these changes by first specifying the \emph{backtracking conditional} $P_B(U^* \mid U)$, a similarity measure between factual exogenous conditions $U$ and counterfactual exogenous conditions $U^*$. This, in turn, induces a joint distribution $P_B(U^*, U) = P_B(U^* \mid U)P(U)$ that allows for the computation of counterfactuals through a `cross-world abduction, marginalization, and prediction' procedure similar to abduction, action, and prediction \cite{von2023backtracking}.
Consider again counterfactual variable $Y^*$ given a factual observation $z$ and --- this time, instead of an intervention --- \emph{counterfactual observation} $X=x^*$. In cross-world abduction, $P_B(U^*, U)$ is updated with evidence $(x^*, z)$ to obtain $P_B(U^*, U \mid x^*, z)$. In marginalization, $U$ is marginalized out to obtain $P_B(U^* \mid x^*, z)$. Finally, prediction computes the probability of $Y^*$ from model $(\mathcal{M}, P_B(U^* \mid x^*, z))$. Observe here that no changes to any causal mechanisms have been made --- we answer the counterfactual query with $\mathcal{M}$ rather than $\mathcal{M}_{x^*}$. Figure~\ref{fig:dags_cf_example} shows a visual representation of this difference for Example~\ref{ex:running_example}, showing the DAGs for an interventional (\ref{fig:dag_regular_cf}) versus backtracking (\ref{fig:dag_backtracking_cf}) approach, and the corresponding exogenous conditions.

\paragraph{Modularity and social categories.} The question of modularity has roots in the philosophy of causal inference and in theories of how cause-and-effect is defined, often discussed in connection with concepts like manipulation \cite{holland_manipulable_1986} or ontological stability \cite{barocas_hardt_narayanan}. We call a variable in a causal model \emph{modular}, or say that it satisfies \emph{modularity}, if (1) its mechanism remains invariant when other mechanisms are subjected to external influences, and (2) if other mechanisms remain invariant when its mechanism is subjected to external influences \cite[p.~60]{Pearl2000CausalityMR}. The notion of modularity arises in interventional counterfactual queries through the fact that intervention $\text{do}(X=x^*)$ leaves all mechanisms other than $\{f_i: V_i \in X\}$ untouched. However, modularity need not be assumed to answer backtracking counterfactual queries because there are no external influences to any mechanisms (no interventions).\\
\indent Questions about modularity also arise naturally in discussions of how to conceptualize counterfactual discrimination criteria and how to incorporate social categories in causal models. In this work, we use the term \emph{social category} to refer to a grouping of people, agnostic to whether or not there are any shared attributes within the group. This makes our framework compatible with groupings that are socially constructed –– by which we mean, created by people in a society. Questioning whether or not variables like social categories can truly be modular (or intervened on at all) has led to questions in the research community of whether causal fairness methods, especially with social categories, are valid or reliable \cite{whats_sex_got_hu_hausmann_2020,Kasirzadeh2021TheUA,causal_fairness_field_guide}. Past work has contended with this issue for treatment choice and policy learning \cite{Bynum2021DisaggregatedIT,Bynum2023Future}. Notably, in this work, the semantics of backtracking counterfactuals allows us to avoid assuming modularity. Our framework thus provides \emph{a possible answer to how we can relax the modularity assumption} and still conceptualize counterfactual discrimination criteria.

\section{Estimating Opportunity}\label{sec:opportunity}

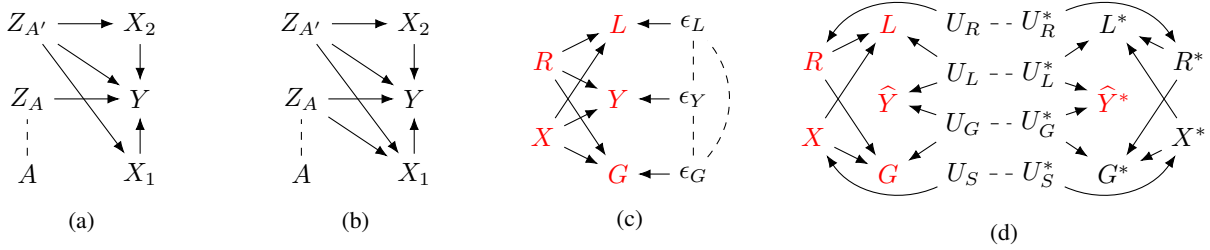
\begin{figure*}
    \centering
    \begin{subfigure}[c]{0.2\textwidth}
    	\centering
        \begin{tikzpicture}
            \node (1) at (-1.5, -1) {$A$};
            \node (2) at (0, -1) {$X_1$};
            \node (3) at (0, 0) {$Y$};
            \node (4) at (-1.5, 0) {$Z_A$};
            \node (7) at (-1.5, 1) {$Z_{A'}$};
            \node (5) at (0, 1) {$X_2$};
        
            \path (2) edge (3);
            \path (4) edge (3);
            \path (7) edge (3);
            \path (7) edge (2);
            \path (7) edge (5);
            \path (5) edge (3);
            \draw[dashed,-] (1) edge (4);
        \end{tikzpicture}
    	 \caption{}
         \label{fig:scenario_1_dag}
    \end{subfigure}
    \begin{subfigure}[c]{0.2\textwidth}
    	\centering
        \begin{tikzpicture}
            \node (1) at (-1.5, -1) {$A$};
            \node (2) at (0, -1) {$X_1$};
            \node (3) at (0, 0) {$Y$};
            \node (4) at (-1.5, 0) {$Z_A$};
            \node (7) at (-1.5, 1) {$Z_{A'}$};
            \node (5) at (0, 1) {$X_2$};
        
            \path (2) edge (3);
            \path (4) edge (3);
            \path (4) edge (2);
            \path (7) edge (3);
            \path (7) edge (2);
            \path (7) edge (5);
            \path (5) edge (3);
            \draw[dashed,-] (1) edge (4);
        	 \end{tikzpicture}
     	 \caption{}
         \label{fig:scenario_2_dag}
    \end{subfigure}
    \begin{subfigure}[c]{0.2\textwidth}
        \centering
        \begin{tikzpicture}
            \node (1) at (-1, 0.5) {\textcolor{red}{$R$}};
            \node (2) at (-1, -0.5) {\textcolor{red}{$X$}};
            \node (3) at (0, 1) {\textcolor{red}{$L$}};
            \node (4) at (0, 0) {\textcolor{red}{$Y$}};
            \node (5) at (0, -1) {\textcolor{red}{$G$}};
            \node (6) at (1, 1) {$\epsilon_L$};
            \node (7) at (1, 0) {$\epsilon_Y$};
            \node (8) at (1, -1) {$\epsilon_G$};
    
            \path (1) edge (3);
            \path (1) edge (4);
            \path (1) edge (5);
    
            \path (2) edge (3);
            \path (2) edge (4);
            \path (2) edge (5);
    
            \path (6) edge (3);
            \path (7) edge (4);
            \path (8) edge (5);
    
            \draw[dashed, -] (6) edge (7);
            \draw[dashed, -] (7) edge (8);
            \draw[dashed, -, bend right=40] (8) edge (6);
        \end{tikzpicture}
        \caption{}
        \label{fig:level_3_dag}
    \end{subfigure}
    \begin{subfigure}[c]{0.35\textwidth}
        \centering
        \begin{tikzpicture}
            \node (1) at (-1, 0.5) {\textcolor{red}{$R$}};
            \node (2) at (-1, -0.5) {\textcolor{red}{$X$}};
            \node (3) at (0, 1) {\textcolor{red}{$L$}};
            \node (4) at (0, 0) {\textcolor{red}{$\widehat{Y}$}};
            \node (5) at (0, -1) {\textcolor{red}{$G$}};
            \node (6) at (1, 0.33) {$U_L$};
            \node (17) at (1, 1) {$U_R$};
            \node (18) at (1, -1) {$U_S$};
            \node (7) at (1, -0.33) {$U_G$};
            
            \node (8) at (4, 0.5) {$R^*$};
            \node (9) at (4, -0.5) {$X^*$};
            \node (10) at (3, 1) {$L^*$};
            \node (11) at (3, 0) {\textcolor{red}{$\widehat{Y}^*$}};
            \node (12) at (3, -1) {$G^*$};
            \node (13) at (2, 0.33) {$U^*_L$};
            \node (15) at (2, 1) {$U^*_R$};
            \node (16) at (2, -1) {$U^*_S$};
            \node (14) at (2, -0.33) {$U^*_G$};

            \path (1) edge (3);
            \path (1) edge (5);
    
            \path (2) edge (3);
            \path (2) edge (5);
    
            \path (6) edge (3);
            \path (6) edge (4);
            \path (7) edge (4);
            \path (7) edge (5);
            \draw[bend right=40] (17) edge (1);
            \draw[bend left=40] (18) edge (2);

            \path (8) edge (10);
            \path (8) edge (12);
    
            \path (9) edge (10);
            \path (9) edge (12);
    
            \path (13) edge (10);
            \path (13) edge (11);
            \path (14) edge (11);
            \path (14) edge (12);
            \draw[bend left=40] (15) edge (8);
            \draw[bend right=40] (16) edge (9);

            \draw[dashed, -] (6) edge (13);
            \draw[dashed, -] (7) edge (14);
            \draw[dashed, -] (17) edge (15);
            \draw[dashed, -] (18) edge (16);
        \end{tikzpicture}
        \caption{}
        \label{fig:level_3_backtracking_dag}
    \end{subfigure}
    \caption{(a) Stylized causal model for a hiring example with age $A$, success measure $Y$, latent variables $Z_{A'} \perp A$ and $Z_A \not\perp A$, and qualifications $X_1, X_2$. (b) An alternate version of (a) where latent variable $Z_A$ also impacts $X_1$. (c) A possible DAG for the law school example, reproduced from \protect\cite{Kusner2017CounterfactualF}. (d) A backtracking counterfactual twin network version of the DAG in (a), now with a Level 3 ICF fair predictor $\widehat{Y}$ as the outcome instead of $Y$.}
    \label{fig:experiment_dags}
\end{figure*}

In this section, we detail how backtracking counterfactuals can serve as a useful formalism to describe \emph{opportunity} in the context of an algorithmic decision.

\paragraph{A backtracking conditional to capture agency.} Recall the backtracking conditional $P_B(U^* \mid U)$ serves as a similarity measure between factual conditions $U$ and counterfactual conditions $U^*$. In other words, $P_B(U^* \mid U)$ tells us which counterfactual worlds to consider (and how heavily to weight them) in an answer to a counterfactual query. In a setting where the agency of an individual is at play, rather than use counterfactuals to estimate what changes are \emph{probable} for an individual to get a different outcome, a more natural first step is to consider what changes are \emph{possible} and then, separately, how difficult those changes are to achieve. For this reason, we propose the following `non-informative' backtracking conditional that focuses on what an individual would or would not be able to change (their mutable vs. immutable variables).

\begin{definition}[Non-informative Backtracking Conditional Distribution]\label{def:noninformative_backtracking_conditional}Consider a partitioning of the variables $U$ into mutable variables $M$ and immutable variables $U \setminus M$.  We define a non-informative  backtracking conditional distribution as 
$P_B(U^* \mid U = (m, n)) = P_M(m) \delta(n).$
Here, $P_M(m)$ is the prior marginal distribution over the variables $m \in M$ and $\delta(n)$ is a Dirac delta distribution (i.e., a point mass) at $n \in U \setminus M$.\footnote{This definition implicitly requires a factorizable original distribution $P(U)$, as is typically the case in an SCM.}
\end{definition}

The non-informative backtracking conditional in Definition~\ref{def:noninformative_backtracking_conditional} makes sense if an individual is capable of changing their mutable covariates an arbitrary amount in order to achieve the goal of changing their outcome. But, to capture other worldviews on agency or encode additional constraints on how much an individual could change their covariates, an arbitrary backtracking conditional $P_B(U^* \mid U)$ can be used, under the restriction that we only consider the conditional $P_B(U^* \mid M=m, N=n, N^*=n)$ in our analysis. In both cases, the underlying concept is that an individual can change only their mutable characteristics. With a means of capturing agency, we now define opportunity through the concept of an \emph{opportunity set}.

\begin{definition}[Opportunity Set]\label{def:opportunity_set}
    An opportunity set $S$ is a set of mutable covariates that we believe capture opportunity in the context of a particular algorithmic decision.
\end{definition}

\begin{figure}
    \centering
     \includegraphics[width=0.48\textwidth]{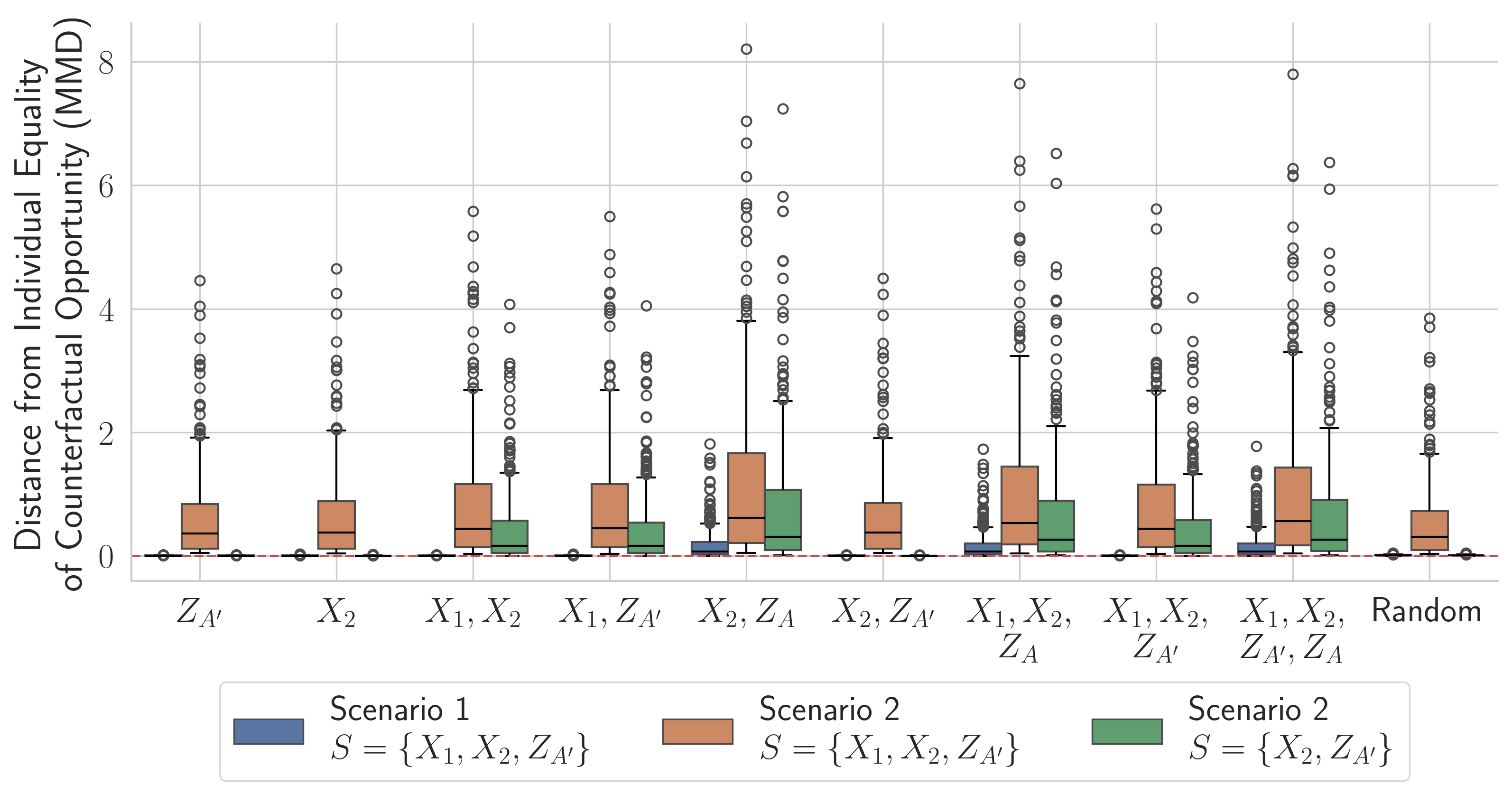}
     \caption{Visualization of how well Individual Equality of Counterfactual Opportunity is satisfied for models using different subsets of covariates across Scenarios \ref{scenario:balanced} and \ref{scenario:unbalanced}, calculated as the absolute energy distance MMD between the two terms of Definition~\ref{def:individual_equal_cf_opportunity}.}
    \label{fig:opportunity_results}
\end{figure}

Returning to Example~\ref{ex:running_example}, we could set $S=\{X\}$ to capture that job-related qualifications $X$ are the mutable covariates that represent an individual's opportunity to get $\widehat{Y} = 0$ or $\widehat{Y} = 1$. In general, given an opportunity set $S$ and a backtracking conditional distribution that describes agency, we can estimate what opportunities different individuals and groups have --- both factually and counterfactually --- to receive particular algorithmic decisions. Across Definitions~\ref{def:individual_cf_opportunity}-\ref{def:group_equal_cf_opportunity}, let $A, X, Y$ represent the protected attributes, remaining attributes, and output of interest, respectively; assume probabilistic causal model $(\mathcal{M}=(U, V, F), P_U)$, where $V \equiv A \cup X$; and consider opportunity set $S \subseteq U \cup V$.

\begin{definition}[Individual-level Counterfactual Opportunity]\label{def:individual_cf_opportunity}
     Consider individual observation $X=x, A=a, \widehat{Y} = y$. The opportunity that this individual has to achieve a counterfactual outcome $\widehat{Y}^* = y^*$ is given by the distribution
     $P(S^*\mid \widehat{Y}^* = y^*, \widehat{Y} = y, X=x, A=a).$
\end{definition}

\begin{definition}[Individual-level Realized Opportunity]\label{def:individual_real_opportunity}
     Consider individual observation $X=x, A=a, \widehat{Y} = y$. The realized opportunity this individual had to achieve outcome $y$ is given by their observed value $s$ of opportunity set $S$.
\end{definition}

\begin{definition}[Group-level Counterfactual Opportunity]\label{def:group_cf_opportunity}
    Consider a group of individuals defined by value $g$ of covariates $G$.\footnote{$G=g$ can be, for example, a setting of protected attributes $A$, a setting of covariates $X$, or both.} The opportunity that this group has to achieve a counterfactual outcome $\widehat{Y}^* = y^*$ is given by the distribution
    $P(S^*\mid \widehat{Y}^* = y^*, G=g).$
\end{definition}

\begin{definition}[Group-level Realized Opportunity]\label{def:group_real_opportunity}
    Consider a group of individuals defined by value $g$ of covariates $G$. The realized opportunity this group had to achieve outcome $y$ is given by the distribution
    $P(S \mid \widehat{Y} = y, G=g).$
\end{definition}

Figure~\ref{fig:main_figure} shows visual examples of Definitions \ref{def:individual_cf_opportunity}~-~\ref{def:group_real_opportunity} for Example~\ref{ex:running_example}. An immediate consequence of these computational descriptions of opportunity is that we can use them to define several new notions of equality of opportunity. 

\begin{definition}[Individual Equality of Counterfactual Opportunity]\label{def:individual_equal_cf_opportunity}
    Consider individual observation $X=x, A=a, \widehat{Y} = y$. Predictor $\widehat{Y}$ satisfies individual equality of counterfactual opportunity for this individual if for all $y^*$,
    \begin{align*}
        P(S^* \mid \widehat{Y}^* = y^*, \widehat{Y} = y, X=x, A=a&) = \\ &P(S^* \mid \widehat{Y}^* = y^*).
    \end{align*}
\end{definition}

In words, Definition~\ref{def:individual_equal_cf_opportunity} captures that an individual's counterfactual opportunities to get an outcome should be the same as the overall population's. We can similarly draw this comparison across those who got the same outcome, capturing equality of \emph{recourse opportunity} with $P(S^* \mid \widehat{Y}^* = y^*, \widehat{Y} = y)$ instead of $P(S^* \mid \widehat{Y}^* = y^*)$.
We could also, by including $X=x$ and/or $A=a'$ as additional conditions on the right hand side, draw a comparison across only those with the same covariates or draw a comparison across groups. In each case, we consider individuals' counterfactual opportunities to get the different possible outcomes. Just as we can compute and equate opportunity for individuals, we can do so for groups.

\begin{definition}[Group-level Equality of Counterfactual Opportunity]\label{def:group_equal_cf_opportunity}
    Predictor $\widehat{Y}$ satisfies group-level equality of counterfactual opportunity if for all $y^*$ and groups $g, g'$,
    $$P(S^* \mid \widehat{Y}^* = y^*,  G=g) = P(S^* \mid \widehat{Y}^* = y^*, G=g').$$
\end{definition}

In words, Definition~\ref{def:group_equal_cf_opportunity} states that each group overall should have the same opportunities to get each outcome. As with Definition~\ref{def:individual_equal_cf_opportunity}, we can add additional conditions to either side of Definition~\ref{def:group_equal_cf_opportunity} for more fine-grained comparisons across groups, e.g., capturing group-level equality of recourse opportunity by adding $\widehat{Y} = y$ to each side.

\section{Describing Effort}\label{sec:effort}

In Section~\ref{sec:opportunity}, we make use of backtracking counterfactuals to capture individuals' \emph{opportunities} to achieve an outcome. In this section, we describe how we can instead capture the \emph{effort required} to achieve an outcome: for example, the cost of recourse. Rather than make assumptions about how much effort an individual may or may not exert in a given situation, we focus instead on how much effort would be required of them, characterized as a difference between two possible states rather than a property of the individual. We can use any appropriate cost function to formalize this difference.

\begin{algorithm}[tb]
    \caption{Backtracking Counterfactual Sampling}
    \label{alg:backtracking_counterfactual_sampling}
    \textbf{Input}: $P(U^* \mid U)$, $\mathcal{M}$, $\{V_i\}_{i=1}^{n}$\\
    \textbf{Parameter}: $n^*$\\
    \textbf{Output}: $\{(U, V, U^*, V^*)_i\}_{i=1}^{n \cdot n^*}$
    \begin{algorithmic}[1]
        \STATE Perform abduction with $\mathcal{M}$ and $\{V_i\}_{i=1}^{n}$ to get $\{U_i\}_{i=1}^{n}$.
        \FOR{$i = 1, \ldots, n^*$}
        \STATE Perform cross-world abduction with $P(U^* \mid U)$ to get one draw of $\{U^*_i\}_{i=1}^{n}$.
        \STATE Perform prediction with $\mathcal{M}$ and $\{U^*_i\}_{i=1}^{n}$ to compute $\{V^*_i\}_{i=1}^{n}$.
        \STATE Append $n$ predicted samples $\{(U, V, U^*, V^*)_i\}_{i=1}^{n}$ from $P(U, V, U^*, V^*)$ to collected data.
        \ENDFOR
        \STATE \textbf{return} samples $\{(U, V, U^*, V^*)_i\}_{i=1}^{n \cdot n^*}$
    \end{algorithmic}
\end{algorithm}

\begin{figure*}
    \centering
    \begin{subfigure}{0.33\textwidth}
        \centering
        \includegraphics[width=\textwidth]{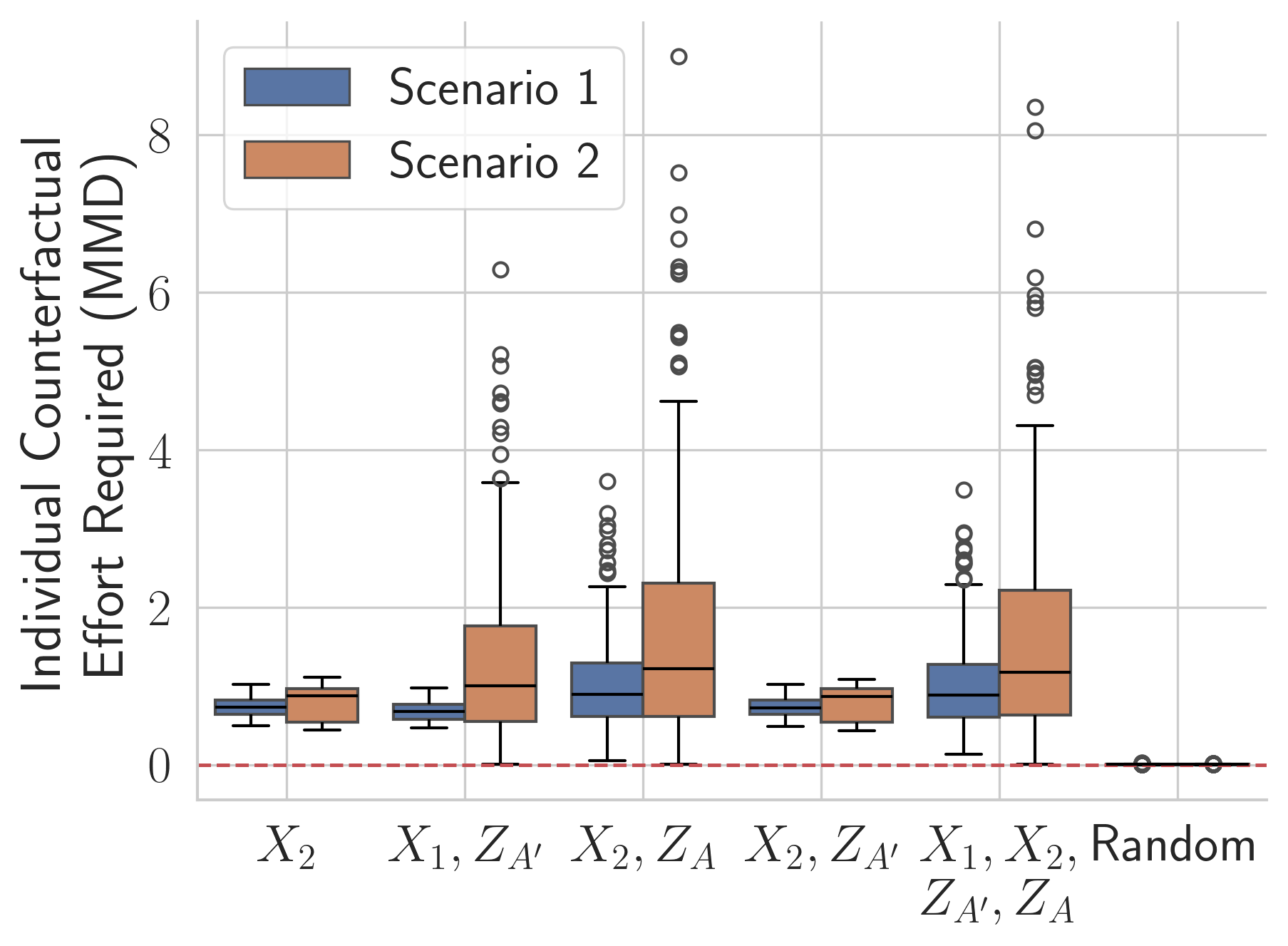}
        \caption{}\label{fig:individual_effort_results}
    \end{subfigure}
    \begin{subfigure}{0.6\textwidth}
        \centering
        \includegraphics[width=\textwidth]{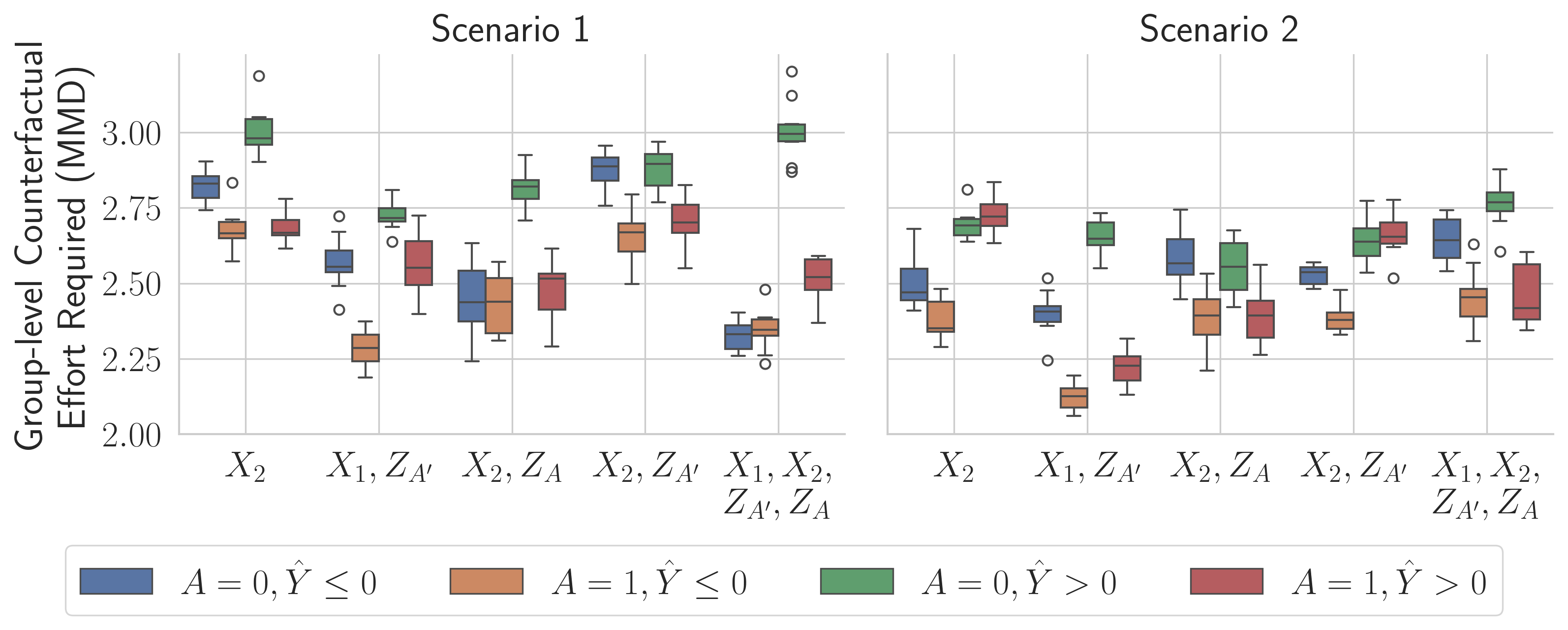}
        \caption{}\label{fig:group_effort_results}
    \end{subfigure}
    \caption{Visualization of (a) Individual Counterfactual Effort Required (Definition~\ref{def:individual_cf_effort}) and (b) Group-level Counterfactual Effort Required (Definition~\ref{def:group_cf_effort}) for models using different subsets of covariates across Scenarios \ref{scenario:balanced} and \ref{scenario:unbalanced}, with energy distance MMD for cost function $\mathcal{L}$.}
    \label{fig:effort_results}
\end{figure*}

\begin{definition}[Individual Counterfactual Effort Required]\label{def:individual_cf_effort}
    Consider individual observation $X=x, A=a, \widehat{Y} = y$ and realized opportunity set $S=s$. The effort required for this individual to achieve counterfactual outcome $\widehat{Y}^* = y^*$, with respect to cost function $\mathcal{L}$, is given by
    $$\mathcal{L} \left(s, P(S^*\mid \widehat{Y}^* = y^*, \widehat{Y} = y, X=x, A=a)\right).$$
\end{definition}
Definition~\ref{def:individual_cf_effort} captures how costly a change from $y$ to $y^*$ is in terms of a distance between what the covariates of interest $S=s$ currently look like compared to what they could look like counterfactually to get the other outcome. The cost function $\mathcal{L}$ represents whatever problem-specific choices we make to account for effort. 

\begin{definition}[Group-level Counterfactual Effort Required]\label{def:group_cf_effort}
    Consider opportunity set $S$ and a group of individuals defined by value $g$ of covariates $G$. The effort required for this group to achieve counterfactual outcome $\widehat{Y}^* = y^*$ given $\widehat{Y}=y$, with respect to cost function $\mathcal{L}$, is defined as
    \begin{align*}
        \textsf{GCE}(g) \equiv \mathcal{L} \left(P(S \mid \vphantom{\widehat{Y}} \right. &\widehat{Y}=y, G=g),\\
        &\left.P(S^* \mid \widehat{Y}^* = y^*, \widehat{Y} = y, G=g)\right).
    \end{align*}
\end{definition}

Definition~\ref{def:group_cf_effort} achieves the same comparison as Definition~\ref{def:individual_cf_effort}, but now for group $g$ instead of an individual. 

As with opportunity, we can equate required effort across individuals or groups as additional counterfactual discrimination criteria. Using the logic of Definition~\ref{def:individual_cf_effort} to capture effort at an individual level, we are left with the question --- present for any individual-level notion of fairness --- of what constitutes a fair baseline for comparison for this individual's effort. Depending on the problem context, Definition~\ref{def:individual_cf_effort} can be used to draw whichever individual-level comparison is desired, and can be computed across individuals as a diagnostic tool to explore which individuals to focus on. At the group level, a natural first comparison to draw is across groups.

\begin{definition}[Group-level Equality of Effort]\label{def:group_equal_cf_effort}
    Predictor $\widehat{Y}$ satisfies group-level equality of effort if $\textsf{GCE}(g) = \textsf{GCE}(g')$ for all groups $g,g'$.
\end{definition}

Several notions of effort and the cost of recourse exist in other literature \cite{Karimi2022ASO}, including work such as \cite{von2022fairness} that frames recourse as a fundamentally causal problem. Even within causal fair recourse literature, analogous notions of recourse cost and required effort are expressed in terms of interventional counterfactuals, and fairness is conceptualized --- particularly at the individual level --- via the traditional paradigm of intervention on $A$. Also unlike other work in recourse, in our case we consider individuals moving both from negative to positive and from positive to negative predictions, more generally capturing a model's behavior. Definitions \ref{def:individual_cf_effort}~-~\ref{def:group_equal_cf_effort} are able to capture classical (asymmetric) recourse if, for example, the cost function $\mathcal{L}$ only accounts for counterfactual changes across the decision boundary from factual outcomes $\widehat{Y} = 0$ to counterfactual $\widehat{Y}^* = 1$ and ignores changes from $\widehat{Y} = 1$ to $\widehat{Y}^* = 0$.

\section{Experiments}

In this section, we demonstrate our criteria computationally on real and simulated data. Using the framework introduced in Sections~\ref{sec:opportunity} and \ref{sec:effort}, a variety of different counterfactual quantities and discrimination criteria can be computed for a given predictor. Exploring all these possibilities computationally is beyond the scope of this short paper: in this section we focus on Definitions~\ref{def:individual_equal_cf_opportunity}, \ref{def:individual_cf_effort}, and \ref{def:group_cf_effort} as representative examples of both opportunity-based and effort-based criteria.

In order to actually compute backtracking counterfactuals, we take a simple approach to estimating the joint distribution $P(U, V, U^*, V^*)$, detailed in Algorithm~\ref{alg:backtracking_counterfactual_sampling}. Although the brute-force approach in Algorithm~\ref{alg:backtracking_counterfactual_sampling} is in theory suitable for any backtracking counterfactual query that conditions on either observed $U, V$ or finite-domain $U^*, V^*$ (given large enough $n$ and $n^*$), it is by nature not scalable. However, the development of efficient or query-first algorithms for sampling backtracking counterfactuals is not our focus here, and the naive approach is suitable for each of our experiments. 

\subsection{Synthetic hiring datasets}\label{sec:hiring_experiments}

To explore a more complex graphical structure than Example~\ref{ex:running_example}, consider the following setup. Figures~\ref{fig:scenario_1_dag} and \ref{fig:scenario_2_dag} show a causal graphical model that pictorially represents how an applicant's qualifications $X_1$ and $X_2$ impact some measure of job success $Y$. In this example, \emph{age group} $A$ is the protected attribute. Everything unmeasured that we believe impacts $X_1, X_2$ and $Y$ is latent, and for simplicity, we assume we can partition the latent space into variables $Z_{A} \not\perp A$, representing age-related (immutable) circumstance, and $Z_{A'} \perp A$, representing (mutable) other factors unrelated to age. In this context, imagine an algorithm used during the hiring process that makes a prediction $\widehat{Y}$ of future success.

\begin{scenario}[Balanced Qualifications]\label{scenario:balanced} 
In this scenario, qualifications $X_1$ and $X_2$ are both balanced across age groups. In other words, the recruited applicant pool has good representation of relevant qualifications across all age groups. Scenario~\ref{scenario:balanced} is generated as follows:
$A \sim \text{Bern}(0.5)$,
$Z_A \sim \mathcal{N}(A/2, 1)$,
$Z_{A'} \sim \mathcal{N}(0, 1)$,
$X_1 \sim \mathcal{N}(Z_{A'}, 1)$,
$X_2 \sim \mathcal{N}(3Z_{A'}, 1)$, and
$Y \sim \mathcal{N}(X_1 + X_2 + 2 Z_A + Z_{A'} - 1, 1)$. The corresponding DAG is shown in Figure~\ref{fig:scenario_1_dag}.\footnote{In this simulation, the impact of $A$ on $Z_{A}$ is not viewed as a causal effect but instead as a convenient way of inducing dependence. This choice does not change any of the results, as no interventions on $A$ are considered.}
\end{scenario}

\begin{scenario}[Unbalanced Qualifications]\label{scenario:unbalanced}
In this scenario, qualification score $X_1$ is not balanced across age groups. This imbalance could arise due to, for example, unequal access to developmental resources beforehand, i.e., age-related circumstance $Z_A$ has impacted qualifications $X_1$. We generate data for Scenario~\ref{scenario:unbalanced} with a small modification to Scenario~\ref{scenario:balanced}:
$A \sim \text{Bern}(0.5)$,
$Z_A \sim \mathcal{N}(A/2, 1)$,
$Z_{A'} \sim \mathcal{N}(0, 1)$,
$X_1 \sim \mathcal{N}(2Z_{A} + Z_{A'}, 1)$,
$X_2 \sim \mathcal{N}(3Z_{A'}, 1)$, and
$Y \sim \mathcal{N}(X_1 + X_2 + 2 Z_A + Z_{A'} - 2, 1)$. The corresponding DAG is shown in Figure~\ref{fig:scenario_2_dag}.
\end{scenario}

We fit several predictors on datasets of size 500 from the above equations, each a linear regression using some subset of the available covariates (excluding $A$). For each predictor, we obtain an estimate of $P(U, V, U^*, V^*)$ via Algorithm~\ref{alg:backtracking_counterfactual_sampling} using a non-informative backtracking conditional following Definition~\ref{def:noninformative_backtracking_conditional}, and estimate the backtracking counterfactual terms of interest from $P(U, V, U^*, V^*)$, where we consider protected groups $A=0$ and $A=1$ with binarized outcome cases $\widehat{Y} > 0$ and $\widehat{Y} \leq 0$. 

Figure~\ref{fig:opportunity_results} shows how well each of these models satisfies Definition~\ref{def:individual_equal_cf_opportunity} across the two scenarios. This is measured as the absolute energy distance maximum mean discrepancy (MMD) between the two terms --- $P(S^* \mid \widehat{Y}^* = y^*, \widehat{Y} = y, X=x, A=a)$ and  $P(S^* \mid \widehat{Y}^* = y^*, A=a')$ ---  \emph{for each individual in the dataset} as a boxplot for each model. When these distances are zero for each individual, Definition~\ref{def:individual_equal_cf_opportunity} is satisfied. Figure~\ref{fig:opportunity_results} shows that in Scenario~\ref{scenario:balanced} with opportunity set $S = \{X_1, X_2, Z_{A'}\}$, a model using any subset of the mutable covariates (i.e., not using age $A$ or age-related circumstance $Z_{A}$) will satisfy Definition~\ref{def:individual_equal_cf_opportunity}. However, in Scenario~\ref{scenario:unbalanced}, with the same opportunity set, no model satisfies Definition~\ref{def:individual_equal_cf_opportunity}. 

This comparison demonstrates an important takeaway. Even a random predictor will not balance an unbalanced predictor (like $X_1$) counterfactually across groups. The subsequent implication of even a random predictor being unfair is that, to achieve fairness, \emph{we would have to change the underlying process that led to the imbalance in $X_1$ in the first place}. This connects to other discussions in, e.g., recourse literature on the need to consider \emph{societal interventions} in addition to modifications of a classifier \cite{von2022fairness}. If, instead, imbalance in $X_1$ is deemed morally acceptable, and $X_1$ is thus not considered part of an individual's opportunity set $S$, then model performance follows a trend much like that for Scenario~\ref{scenario:balanced}, where a model that doesn't make use of $Z_{A}$ nor its now-descendant $X_1$ can easily satisfy Definition~\ref{def:individual_equal_cf_opportunity}. 

Figure~\ref{fig:effort_results} shows individual and group-level counterfactual effort required (Definitions \ref{def:individual_cf_effort} and \ref{def:group_cf_effort}) for the same datasets given opportunity set $S = \{X_1, X_2, Z_{A'}\}$ and MMD as cost function $\mathcal{L}$. Figure~\ref{fig:individual_effort_results} shows how much effort is required of each individual in the dataset to get the opposite of their current outcome, showing wide variation in the cost of changing an outcome across different individuals, as well as more costly changes in Scenario~\ref{scenario:unbalanced} than in Scenario~\ref{scenario:balanced}. Figure~\ref{fig:group_effort_results} shows how much effort is required instead for each group ($A=0$ and $A=1$) to flip their outcome starting from negative ($\widehat{Y} \leq 0$) or starting from positive ($\widehat{Y} > 0$), this time across 10 runs to account for variation in estimating MMD.\footnote{In this case, we have one sole computation per model rather than one for each individual in the dataset.} This figure demonstrates that even for the same model, group-level takeaways can often be the opposite of individual-level takeaways: at the group-level, Scenario~\ref{scenario:balanced} is often more costly than Scenario~\ref{scenario:unbalanced}. Figure~\ref{fig:group_effort_results} also shows that changing outcomes is typically harder for group $A=0$ than for group $A=1$, across both scenarios. 

These quantities --- and our takeaways from using them ---
are immediately applicable in real data settings, like the one discussed in Section~\ref{sec:law_school_experiments}.

\subsection{Law school dataset}\label{sec:law_school_experiments}

\begin{figure}
    \centering
    \includegraphics[width=0.35\textwidth]{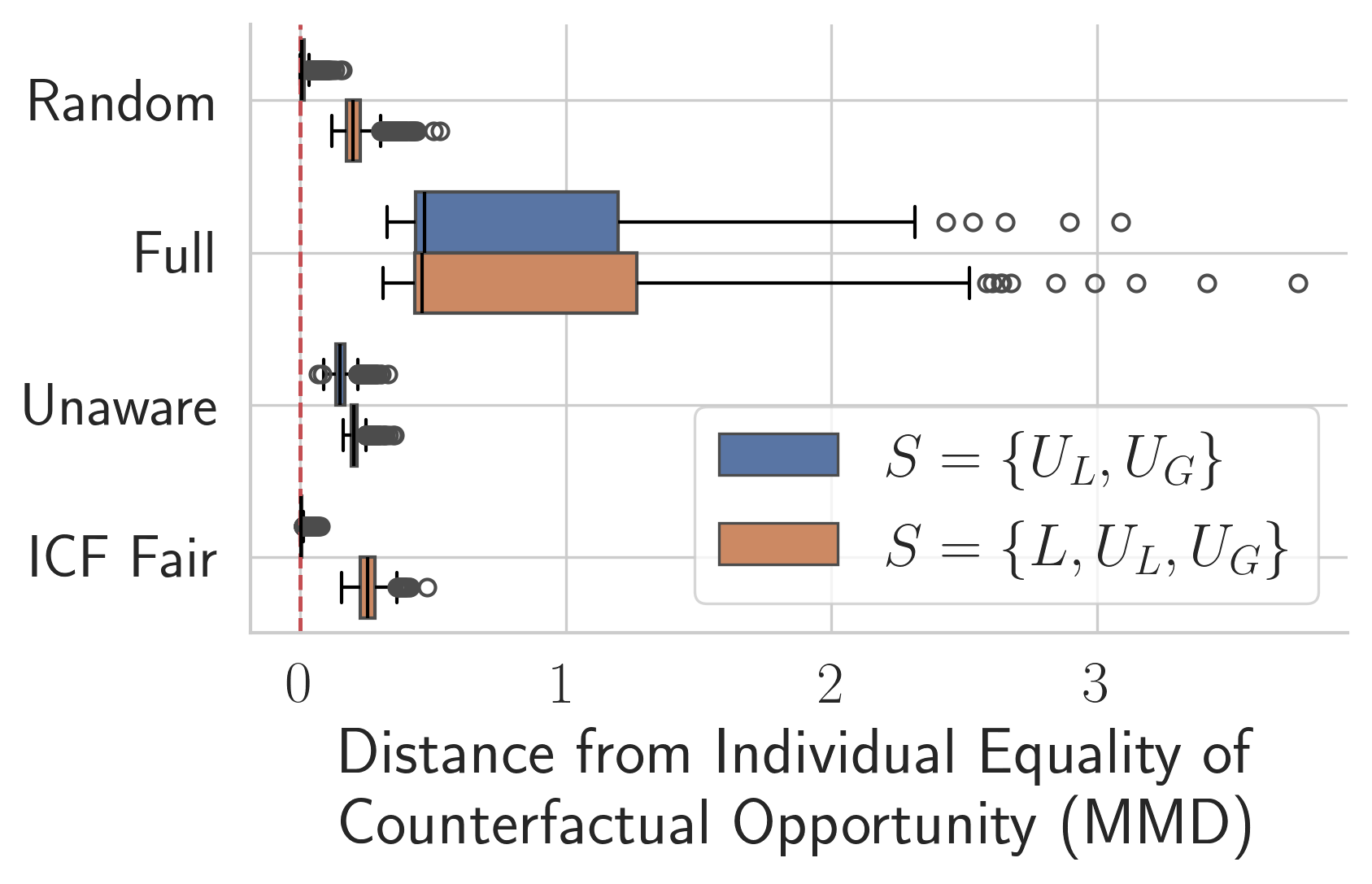}
    \caption{Visualization of Individual Equality of Counterfactual Opportunity on the law school dataset for four predictors: a random model (Random), a model using all available covariates (Full), an `unaware' model using only $L, G$ (Unaware), and a model satisfying interventional counterfactual fairness (ICF Fair). (a) Opportunity set $S = \{U_L, U_G\}$. (b) Opportunity set $S=\{L, U_L, U_G\}$.}
    \label{fig:law_school_results}
\end{figure}

Used to illustrate counterfactual fairness in \cite{Kusner2017CounterfactualF}, the law school dataset from \cite{Wightman1998LSACNL} contains information on law students’ demographics and academic performance compiled from a 1998 Law School Admission Council survey. \citeauthor{Kusner2017CounterfactualF} imagine a task of predicting the average grade of prospective students in their first year of law school in conjunction with the causal graph depicted in Figure~\ref{fig:level_3_dag}, where variables $G, L, R, X, Y$ represent grade point average, entrance exam (LSAT) score, racial category, sex, and first-year average grade, respectively. Based on the structure of the model from \cite{Kusner2017CounterfactualF}, we have SCM $L = f_L(R, X) + \epsilon_L$; $G = f_G(R, X) + \epsilon_G$; $Y \sim \text{Bern}(p=\text{expit}(f_Y(R, X)))$; with linear functions $f_L, f_G, f_Y$ and normally distributed $\epsilon_L, \epsilon_G$.  For this dataset, focusing on $R$ binarized to majority/minority and $Y$ to high/low, Definition~\ref{def:individual_equal_cf_opportunity} compares $P(S^* \mid \widehat{Y}^*=y^*, R=r, X=x, G=g, L=\ell, \widehat{Y}=y)$ to $P(S^* \mid \widehat{Y}^*=y^*, R=r')$ for each individual.

One method to train a model that satisfies traditional, interventional counterfactual fairness (ICF) is to postulate a fully deterministic model with latent variables and use only the latent variables as inputs to the predictor. A `Level 3' ICF fair predictor $\widehat{Y}$ fits separate regressions for $L$ and $G$ and uses residual estimates of $\epsilon_L, \epsilon_G$ to predict $Y$. Figure~\ref{fig:law_school_results} shows MMD distances (in the same manner as Figure~\ref{fig:opportunity_results}) for this model and several others on a random sample of 5000 observations, with two different choices of opportunity sets. Distances are shown for the same sample using either $S = \{U_L, U_G\}$ or $S = \{L, U_L, U_G\}$. In the latter case, we, again, see that none of the predictors satisfy Definition~\ref{def:individual_equal_cf_opportunity}. This figure demonstrates that our choices about what imbalance is morally permissible or not (in this case, with LSAT scores) have relevance for finding a fair model not only in simulations but also in practice. These results also demonstrate empirically how the counterfactual discrimination criteria we introduce here not only capture something philosophically different from interventional criteria, but can also have different implications for the models we use.

\section{Conclusion}

In this work, we have introduced, to our knowledge, the first counterfactual-based fairness criteria that depart from the need to consider intervention on legally-protected characteristics, while still allowing us to consider fairness at an individual or a group level.
We believe this paper can serve as a first step in enabling a new approach to applying counterfactual reasoning to demographic data for socially-relevant concepts like discrimination, charting a course for significant future exploration and technical development. 

\bibliographystyle{named}
\bibliography{references}

\end{document}